\documentclass[letterpaper, 10pt, conference]{styles/ieeeconf}
\IEEEoverridecommandlockouts %
\usepackage{cite}
\usepackage{amsmath,amssymb,amsfonts}
\usepackage{graphicx}
\usepackage{xcolor}
\usepackage{booktabs}
\usepackage{url}
\usepackage{tikz}
\usetikzlibrary{positioning,arrows.meta,fit,calc}

\newif\ifanon

\title{\LARGE \bf GLASS: Architecture-Tuned, Composable, Device-Side\\Linear Algebra for Edge Robotics and Beyond}

\ifanon
    \author{Anonymous Author(s)$^{1}$%
    \thanks{Anonymous funding sources supported this project.}
    \thanks{$^{1}$ Anonymous Institution {\tt\footnotesize anonymous@anonymous}}%
    }
\else
    \author{Brian Plancher$^{1}$%
    \thanks{This project was supported by the National Science Foundation (Award 2411369) and the Toyota Research Institute. Any opinions, findings, conclusions, or recommendations expressed in this material are those of the authors and do not necessarily reflect those of the funding organizations.}%
    \thanks{$^{1}$ Dartmouth College {\tt\footnotesize plancher@dartmouth.edu}}%
    }
\fi

\begin{document}
\maketitle
\thispagestyle{empty}
\pagestyle{empty}

\begin{abstract}
    GPU robotics lacks the reusable numerical infrastructure of mature CPU stacks, instead relying on compiler frameworks that introduce overhead or repeatedly reimplementing numerical libraries.
To address this, we introduce GLASS (GPU Linear Algebra Simple Subroutines), a header-only CUDA C++ library that provides thread-, warp-, block-, and NVIDIA-backed implementations of robotics-scale linear algebra and geometric computations under one composable device API.
GLASS treats implementation choice, execution scope, and launch packing as architecture-specific placement decisions determined by offline measurement and resolved statically at compile time.
This is critical as the best and worst placements differ by a median of 4.9$\times$ (max 81$\times$), with 145 of 396 recommended placements changing between a Jetson AGX Orin and an RTX~5090, and 162 of 396 versus an AGX Xavier.
These stakes are highest at the edge as GLASS's advantage over the best of PyTorch and JAX is as much as 73$\times$ on the Orin versus 12$\times$ on the RTX~5090.
GLASS is released open source with independent numerical oracles and source-bound local-GPU test attestation. 
Finally, integrating GLASS with published robotics systems both exposed a pre-existing numerical bug and improved embedded runtimes by up to 1.5$\times$.
\end{abstract}

\section{Introduction} \label{sec:intro}
Robotics computation is rapidly moving to the GPU, from perception, mapping, localization, and learning to simulation, dynamics, planning, and control~\cite{lang2019pointpillars,freeman2021brax,plancher2022grid,sundaralingam2023curobo,kaufmann2023champion,millane2024nvblox,koide2024megaparticles,adabag2024mpcgpu,vlahov2024mppi,jeon2024cusadi,amatucci2025primal}.
Unfortunately, unlike on the CPU, where robotics applications achieve high performance through the use of libraries that concentrate software optimization, validation, and maintenance into reusable shared infrastructure (e.g.,  Eigen and BLASFEO for linear algebra~\cite{eigen2010,frison2018blasfeo}, Pinocchio and GTSAM for geometric, kinematic, dynamic, and factor-graph kernels~\cite{carpentier2019pinocchio,dellaert2012gtsam}), 
GPU edge robotics lacks such a comparable layer.

This persists because the small, structured operations embedded inside larger robotics kernels still require choosing how computation should map onto GPU threads, warps, and blocks, and whether vendor-backed routines are applicable.
As such, it is unclear whether to leverage high-level compiler frameworks like JAX and PyTorch~\cite{frostig2018jax,ansel2024pytorch2}, use NVIDIA host libraries such as cuBLAS and cuSOLVER~\cite{nvidia2025cublas, nvidia2025cusolver}, integrate device-callable functionality through CUB, CUTLASS, cuBLASDx, and cuSOLVERDx~\cite{nvidia2025cub,nvidia2025cutlass,nvidia2025mathdx}, or repeatedly build and tune bespoke low-level code.
As shown in prior work, and confirmed in this work, no one of these choices is universally best. As such, robotics systems still rely on a mix of all of these methods, incurring overheads and creating duplicate low-level code~\cite{freeman2021brax,plancher2022grid,sundaralingam2023curobo,adabag2024mpcgpu,vlahov2024mppi,jeon2024cusadi,amatucci2025primal}.

Unfortunately, the cost of this duplication is growing as AI coding tools make new implementations much cheaper to produce but not cheaper to validate or maintain~\cite{wang2026maintaincoder,ghammam2026ai,lange2025sakana}. This is particularly problematic for GPU code, which is prone to subtle errors~\cite{yu2005racetrack,betts2012gpuverify}, and for which continuous testing is often too expensive.
Thus, a shared GPU numerical software layer could help amortize optimization, validation, and maintenance across human- and agent-written software.

As such, we introduce \textbf{GLASS (GPU Linear Algebra Simple Subroutines)}, an open source, header-only CUDA C++ library of composable device-side primitives for robotics-scale linear algebra and geometric computation. 
GLASS unifies, under one API, thread-, warp-, and block-scoped implementations alongside NVIDIA device library wrappers. 
GLASS uses offline measurement to determine architecture-specific dispatch and launch placements, resolved statically at compile time.
This is critical as the best and worst placements differ by a median of 4.9$\times$ (max 81$\times$), with 145 of 396 recommended placements changing between a Jetson AGX Orin and an RTX~5090, and 162 of 396 versus an AGX Xavier.
\ifanon
\else
These stakes are highest at the edge as GLASS's advantage over the best of PyTorch and JAX is as much as 73$\times$ on the Orin versus 12$\times$ on the RTX~5090.
\fi

In short, GLASS transforms the numerical layer edge GPU robotics repeatedly rebuilds, or incurs overheads on, into tuned and tested infrastructure. 
Our contributions are:
\begin{enumerate}
\item \textbf{Composable GPU numerical infrastructure.}
GLASS provides one device-side API spanning thread-, warp-, block-, and NVIDIA-backed implementations of dense and structured linear algebra and robotics geometry.

\item \textbf{Measured execution placement.}
GLASS treats implementation choice, execution scope, and launch packing as offline-measured, architecture-specific optimization decisions, leading to both speedups and changing recommendations between hardware targets.

\item \textbf{Robotics-aware composition.}
Composable primitives keep intermediates on device and enable domain-specific fusion, itself a measured placement choice, providing as much as 1.75--2.9$\times$ speedups.

\item \textbf{Validated reuse in robotic systems.}
GLASS provides independent oracles and source-bound GPU test attestation. Integrating GLASS with published robotics systems exposed a pre-existing numerical bug and improved embedded runtimes by up to 1.5$\times$.
\end{enumerate}

\ifanon
\else
We release GLASS open source to benefit the wider community at:
{\texttt{\url{github.com/A2R-Lab/GLASS}}}
\fi

\section{Background and Related Work} \label{sec:related}
\subsection{CPU and GPU Numerical Libraries}

Robotics has long benefited from shared numerical infrastructure on the CPU. Eigen and BLASFEO provide general and small-problem linear algebra~\cite{eigen2010,frison2018blasfeo}, while Pinocchio and GTSAM provide reusable geometric, dynamic, and estimation machinery~\cite{carpentier2019pinocchio,dellaert2012gtsam}.
BLASFEO is particularly relevant because it treats the small-to-medium matrices of embedded optimization as a distinct performance regime~\cite{frison2018blasfeo}. LIBXSMM similarly demonstrates substantial specialization headroom for small matrix multiplication~\cite{heinecke2016libxsmm}.
GLASS targets an analogous reusable numerical layer for \emph{GPU-resident} robotics computation.

The GPU ecosystem already provides strong numerical libraries at several abstraction levels (Table~\ref{tab:landscape}).
Host-dispatched cuBLAS/cuSOLVER, MAGMA, and KBLAS provide optimized dense and batched linear algebra~\cite{haidar2015magma,abdelfattah2016kblas,charara2019batched,abdelfattah2021bblas}.
Ginkgo similarly fuses complete batched iterative and structured solvers into GPU kernels~\cite{anzt2022ginkgo}.
At a higher level, JAX and PyTorch expose compiled, differentiable batched linear algebra~\cite{frostig2018jax,ansel2024pytorch2}, but their operations remain host-dispatched behind framework and kernel-launch boundaries and so cannot be embedded inside application kernels.
More recent NVIDIA libraries move more of this computation inside user kernels.
MathDx provides device-callable numerical routines, CUB provides warp- and block-level collectives, and CUTLASS provides hierarchical matrix kernels with profiler-supported tuning~\cite{nvidia2025mathdx,nvidia2025cub,nvidia2025cutlass}.
Kokkos Kernels provides composable serial, team, and team-vector batched routines across architectures~\cite{rajamanickam2021kokkos}, while Eigen can execute a subset of its operations serially in CUDA threads~\cite{eigen2010}.

GLASS does not seek to replace such device libraries, but rather integrates them as candidate backends alongside native implementations under a single API that can be automatically tuned per target architecture to jointly optimize execution scope and launch packing, saving the optimal setup as a reusable configuration.

\subsection{Performance Portability and Validation}

Numerical software has long used measurement to adapt implementations to a target machine. 
ATLAS benchmarks alternative kernels at installation time~\cite{whaley2001atlas}, FFTW measures candidate execution plans~\cite{frigo2005fftw}, and OSKI selects sparse-kernel implementations from offline measurements~\cite{vuduc2005oski}.
A different approach is to generate or search for specialized code, as in Halide, TVM/Ansor, Triton, and Exo~\cite{raganKelley2013halide,chen2018tvm,zheng2020ansor,tillet2019triton,ikarashi2022exo}.
GLASS follows the measurement-based approach, applying it to execution placement inside GPU kernels. %

Reusable numerical infrastructure must also preserve correctness across these implementations. 
GPUVerify targets synchronization and race errors~\cite{betts2012gpuverify}, while recent work on generated GPU kernels shows that apparent performance can depend strongly on the correctness oracle used for validation~\cite{ouyang2025kernelbench,lange2025sakana,zhang2026kbverified,sarkar2026illusion}. 
GLASS therefore pairs its implementations with independent numerical oracles. Its engineering workflow can also carry source-bound, low-cost, local-GPU test results into CPU-only Continuous Integration (CI). %

\begin{table}[t]
\caption{Representative GPU numerical libraries and the capabilities. Parentheses denote partial support (e.g., a subset of functions, profiler- or heuristic-guided choices).}
\label{tab:landscape}
\centering
\footnotesize
\setlength{\tabcolsep}{2pt}
\renewcommand{\arraystretch}{1.10}
\begin{tabular}{@{}p{2.7cm}p{1.7cm}ccc@{}}
\toprule
& & \textbf{LA + Fact.} & \textbf{Robotics} & \textbf{Guided} \\
\textbf{Library} & \textbf{Scope} & \textbf{+ Solves} &
\textbf{+ fused} & \textbf{selection} \\
\midrule
JAX, PyTorch & host$\to$batch & \checkmark & --- & (\checkmark) \\
cuBLAS/cuSOLVER & host$\to$batch & \checkmark & --- & --- \\
MAGMA, KBLAS & host$\to$batch & \checkmark & --- & (\checkmark) \\
cuBLASDx/\allowbreak cuSOLVERDx & thread/\allowbreak block & \checkmark & --- & --- \\
CUB, CUTLASS & thread/\allowbreak warp/\allowbreak block & (\checkmark) & --- & (\checkmark) \\
Kokkos Kernels & thread/block & (\checkmark) & --- & --- \\
Eigen (device) & thread & (\checkmark) & --- & --- \\
\midrule
\textbf{GLASS} & \textbf{thread/\allowbreak warp/\allowbreak block} & \checkmark &
\checkmark & \checkmark \\
\bottomrule
\end{tabular}
\end{table}

\section{Design} \label{sec:design}
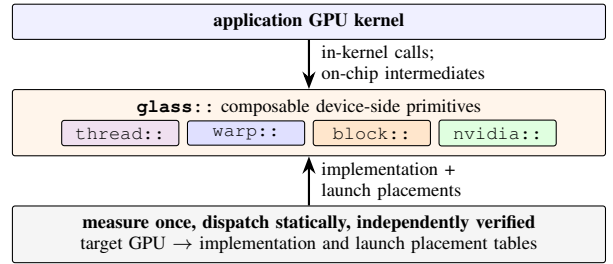
\begin{figure}[t]
\centering
\begin{tikzpicture}[
  font=\scriptsize,
  node distance=0.22cm,
  box/.style={
    draw,
    rounded corners=1pt,
    inner sep=3.5pt,
    align=center,
    text width=0.88\columnwidth
  },
  tier/.style={
    draw,
    rounded corners=1pt,
    inner sep=2.5pt,
    align=center,
    font=\scriptsize\ttfamily,
    minimum width=1.55cm
  },
  flow/.style={-{Stealth}, thick},
]

\node[box, fill=blue!6] (app)
  {\textbf{application GPU kernel}};

\node[box, fill=orange!8, below=0.65cm of app] (api)
  {\textbf{\ttfamily glass::} \normalfont composable device-side primitives\\[1.5pt]
   \begin{tikzpicture}[node distance=0.10cm]
     \node[tier, fill=violet!12] (t) {thread::};
     \node[tier, fill=blue!12, right=of t] (w) {warp::};
     \node[tier, fill=orange!20, right=of w] (b) {block::};
     \node[tier, fill=green!12, right=of b] (n) {nvidia::};
   \end{tikzpicture}};

\node[box, fill=black!4, below=0.65cm of api] (tune)
  {\textbf{measure once, dispatch statically, independently verified}\\
   target GPU $\rightarrow$ implementation and launch placement tables};

\draw[flow] (app) -- node[right=1pt, align=left]
  {in-kernel calls;\\on-chip intermediates} (api);

\draw[flow] (tune) -- node[right=1pt, align=left]
  {implementation +\\launch placements} (api);

\end{tikzpicture}
\vspace{-5pt}
\caption{GLASS at a glance. Application kernels call one device-side API spanning thread-, warp-, block-, and NVIDIA-backed implementations. 
Offline measurement produces architecture-specific compile-time dispatch and launch placements. Independent numerical tests validate every implementation
separately from performance measurement.}
\label{fig:overview}
\vspace{-10pt}
\end{figure}

GLASS (Fig.~\ref{fig:overview}) is a header-only CUDA C++ library of composable device-side primitives for robotics-scale linear algebra and geometric computation designed for the small- to medium-sized batched computations common in robotics. Application kernels call these primitives directly, allowing numerical operations to remain inside the surrounding GPU computation and avoiding additional host calls, kernel launches, or data marshalling. 

\subsection{Programming Model and Execution Interfaces}
\label{sec:interfaces}

In the canonical interface, one CUDA block owns one independent numerical problem and its threads cooperate over that problem, with alternative interfaces also available. %
In particular, \texttt{glass::block::} uses dependency-free cooperative SIMT, \texttt{glass::warp::} assigns one problem to a warp, and \texttt{glass::thread::} assigns one problem to a thread.
GLASS also exposes optional NVIDIA-backed implementations through \texttt{glass::nvidia::*::}, where \texttt{*} is \texttt{block}, \texttt{warp}, or \texttt{thread}. These interfaces wrap CUB collectives, cuBLASDx matrix operations, and cuSOLVERDx factorizations and solves where available~\cite{nvidia2025cub,nvidia2025mathdx}.
Importantly, all of these interfaces share the same mathematical conventions and data layouts while making execution scope and optional dependencies explicit.

Choosing among these implementations is architecture and application dependent and would quickly become a challenge for users. 
GLASS therefore ships with autotuning tools that benchmark the library's performance offline on a target GPU and store the resulting choices in architecture-specific tables.
The selected implementation body is then folded through an architecture-specific \texttt{constexpr} table at compile time, avoiding host dispatch and runtime overheads.

GLASS provides two ways to use these measurements. %
A bare call such as \texttt{glass::posv} keeps the standard block-level calling contract but automatically selects the best compatible implementation body from the measured table. Changing from one problem per block to one per warp or thread also changes application launch geometry and indexing, so GLASS cannot make that transformation implicitly. 
Thus, we also provide a compile-time advisor that reports the recommended execution scope and launch packing, allowing the application to adopt the measured mapping when desired. 
For example, on the Orin, for \texttt{fp32} POSV, the \texttt{glass::recommend} API places \texttt{glass::thread::} at $N{=}8$ but \texttt{glass::nvidia::block::} at $N{=}32$.

Finally, GLASS deliberately targets numerical problems small enough to be owned by one CUDA block. Such problems are common in edge robotic applications, and this is the regime in which embedding a routine inside an application kernel avoids costly overheads. Larger cross-block dense problems are generally better served by conventional host-dispatched libraries. Within the targeted regime, each implementation is instantiated only where its compile-time size and resource requirements are feasible.

\subsection{Numerical and Robotics Primitives}
\label{sec:operators}

GLASS's linear algebra surface covers much of the BLAS and LAPACK functionality needed by robotics applications, including reductions, BLAS L1--L3 operations, triangular and symmetric updates, factorizations, and linear solves. 
GLASS also includes structured routines such as block-tridiagonal matrix-vector products and direct and iterative solvers commonly used in trajectory optimization~\cite{adabag2024mpcgpu,frison2018blasfeo}.
On top of this general numerical layer, GLASS adds operations that recur throughout robotics software. These include spatial algebra, SO(3), SE(3), and quaternion maps and Jacobians, pose errors and retractions, projections, and related geometric and optimization primitives~\cite{carpentier2019pinocchio,plancher2022grid}. 
Sharing these implementations is useful for correctness as well as performance because robotics libraries often differ in twist ordering, quaternion layout, perturbation conventions, and small-angle behavior.
GLASS fixes these conventions as part of each operation's interface and tests them against independent references or defining identities.

Because these primitives share an in-kernel interface, they can also be composed without introducing new kernel boundaries. Higher-level operations can therefore reuse the same library calls while retaining intermediate data on chip, providing performance gains. For example, the LQR feedback gain, 
$K = (R + B^\top P B)^{-1} B^\top P A$,
can be assembled from general matrix and solve primitives or optimized with a fused \texttt{riccati\_gain} implementation (Sec.~\ref{sec:robotics_eval}). As such, we include a number of fused operations in GLASS.

\subsection{Validation and Continuous Testing}
\label{sec:correctness}

Because GLASS may select different implementations across execution scopes and GPU architectures, numerical correctness must hold independently of the chosen implementation. We therefore validate conventional linear algebra against independent NumPy and SciPy references~\cite{harris2020numpy,virtanen2020scipy}, and robotics operations against Pinocchio where applicable~\cite{carpentier2019pinocchio}.
Operations without a direct reference are checked using defining identities, factorization residuals, algebraic equivalences, or finite-difference derivatives. 
Tests also cover numerical scale, conditioning, layouts, and execution scopes where these form part of the documented contract.

Continuous testing of this library presents a practical problem because GPU runners in hosted CI are comparatively expensive. 
We therefore developed a lightweight \texttt{pytest} plugin that runs GPU tests on available local hardware, records the exact source tree and test outcomes, and signs this evidence for later verification in CPU-only CI. 
Subsequent CI runs can reject stale or source-mismatched results without rerunning the GPU test suite. 
We note that these receipts are engineering attestations by their signers, not proof that a claimed GPU executed the tests. However, we include mechanisms to reduce the viable signer list to a trusted subset of developers, e.g., for major releases, to provide higher levels of confidence in the testing regime.
\ifanon
\else
We release this testing tool open source alongside GLASS at:\\\url{github.com/A2R-Lab/pytest-gpu-proof}\\and via: \texttt{pip install pytest-gpu-proof}.
\fi

\section{Evaluation} \label{sec:eval}
\begin{figure*}[t]
\centering
\includegraphics[width=\textwidth]
{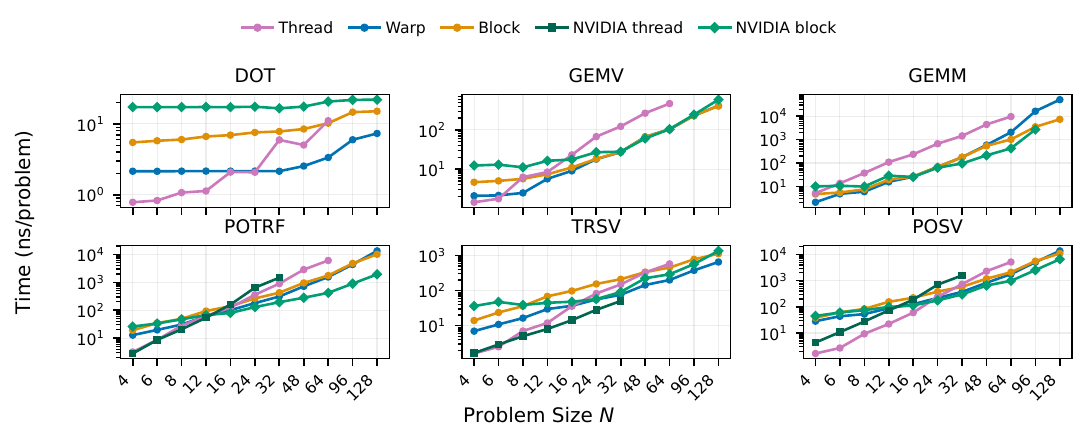}
\vspace{-25pt}
\caption{Measured \texttt{fp32} kernel timings on the Jetson AGX Orin at $B=8192$. Each panel reports per-problem time as problem size varies. Native GLASS Thread, Warp, and Block implementations are measured alongside NVIDIA device paths where supported. Sampled problem sizes are spaced uniformly to expose the small-$N$ regime. Crossings show why no single execution scope provides the best implementation and why placement matters.}
\label{fig:device_vendor}
\vspace{-12pt}
\end{figure*}

Five key questions anchor our evaluation. First, what does GLASS provide beyond existing device-side numerical libraries? Second, how strongly do placements depend on GPU architecture? Third, how does this execution model compare with conventional host-batched libraries and compiler frameworks? Fourth, do the same choices benefit robotics-specific operations and compositions? Finally, does replacing bespoke code with GLASS improve real robotics software?

\subsection{Methodology}
\label{sec:methodology_eval}

Our evaluation spans the regimes exposed by GPU robotics. GPU simulation and sampling-based methods can evaluate batches containing hundreds to thousands of numerical problems~\cite{vlahov2024mppi,makoviychuk2021isaacgym,jeon2024cusadi}, trajectory optimization commonly exposes batch sizes in the tens to hundreds~\cite{du2026gato}, and high-rate control also makes the batch-one latency regime important~\cite{adabag2024mpcgpu}. We thus evaluate batch sizes from $B=1$ to $B=8192$ and problem sizes from $N=4$ to $N=128$, ensuring that we span robotics-scale, low-latency single-problem execution, through heavily batched GPU workloads.

We use the embedded Jetson AGX Orin (sm\_87, CUDA~13.2) as our primary test platform. A Jetson AGX Xavier (sm\_72, CUDA~11.4) provides an older-generation edge comparison, and a desktop RTX~5090 (sm\_120, CUDA~13.2) provides a high-performance ablation that tests whether the same measured-placement design scales up. 
Benchmarks preheat the GPU, use repeated measurements, and validate implementations independently from timing.\footnote{Placement benchmarks measure wall-clock time around back-to-back asynchronous launches followed by one synchronization, while case-study A/B experiments use CUDA events. Throughput comparisons exclude host API overhead, favoring the host-dispatched baselines, while synchronized batch-one measurements separately capture call latency.}

Placement sweeps cover six operations, eleven problem sizes, three batch regimes, and two scalar types (\texttt{fp32}, \texttt{fp64}), for 396 cells per architecture. Host comparisons cover GEMM, POTRF, and POSV over nine sizes, seven batch sizes, and both scalar types, for 378 cells, and use only GLASS's dependency-free native tiers as candidates.\footnote{Destructive POTRF, TRSV, and POSV benchmarks give every supported native and NVIDIA execution plan a fresh valid input per launch, randomize plan order within paired rounds, and retain all raw samples.} We note that due to structural limitations, the NVIDIA backend supports only 357 of 396 (195 of 198 \texttt{fp32}) cells on the Orin, 354 (192 \texttt{fp32}) on the RTX~5090, and 0 on the Xavier.\footnote{Its CUDA~11.4 toolchain predates the device-callable MathDx libraries, requiring it to always fall back to native GLASS implementations.} We call these the \emph{comparable} cells. Finally, NVIDIA placements are chosen when they are $>5$\% faster, the \emph{dispatch margin}, to account for measurement noise.

\subsection{Performance Against Device-Side Libraries}
\label{sec:device_eval}

As GLASS is designed to execute inside application GPU kernels, its closest comparisons are other device-side numerical implementations.
Figure~\ref{fig:device_vendor} shows the placement measurement sweep for $B=8192$ for \texttt{fp32} operations on the Jetson AGX Orin. The crossing curves expose the value of retaining several execution scopes rather than standardizing on one implementation family, with the best and worst placements differing by a median of 4.9$\times$ and up to 81$\times$. In particular, native GLASS wins 99 of the 195 comparable \texttt{fp32} cells (51\%), most often for dot, GEMV, GEMM, and triangular solves. These wins have a geometric mean of $2.1\times$, with 39 exceeding $2\times$, and reaching $22.1\times$ in the extreme case where thread-packed execution replaces a block-wide reduction at tiny $N$. NVIDIA implementations are selected or lie within the dispatch margin in 96 of 195 comparable cells (49\%), dominated by factorization operations. On the workstation RTX~5090 the NVIDIA-backed share rises to 64\% with a maximum native win of $3.9\times$ on its 192 comparable cells.

\begin{figure}[t]
\centering
\includegraphics[width=\linewidth]{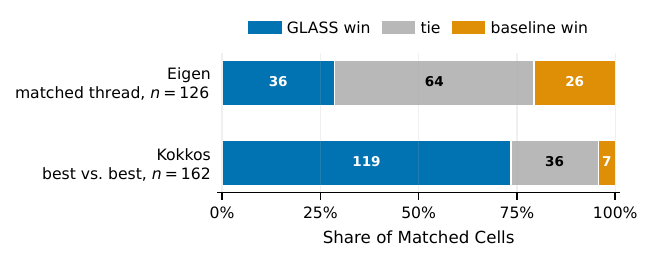}
\vspace{-25pt}
\caption{Other device-side comparisons over limited matched supported operations on the Jetson AGX Orin, confirming GLASS's advantage due to its broader set of execution placements and numerical primitives.}
\label{fig:device_baselines}
\vspace{-12pt}
\end{figure}

Other device-side library baselines reinforce this result (Figure~\ref{fig:device_baselines}).
As Eigen can only be embedded into individual threads, we compare it to \texttt{glass::thread::} and find that GLASS wins 36 of 126 cells, Eigen wins 26, the remaining 64 tie, and the median runtime ratio is 1.00$\times$.
Kokkos Kernels provides a broader composable comparison spanning serial and cooperative execution. Here, GLASS wins 119 of 162 cells, Kokkos wins 7, and the other 36 tie.

Taken together, these benchmarks show that GLASS's advantage comes from its broader set of placements and primitives, and this advantage is most important for deployable robotics-scale edge devices.

\begin{figure*}[!t]
\centering
\includegraphics[width=\textwidth]{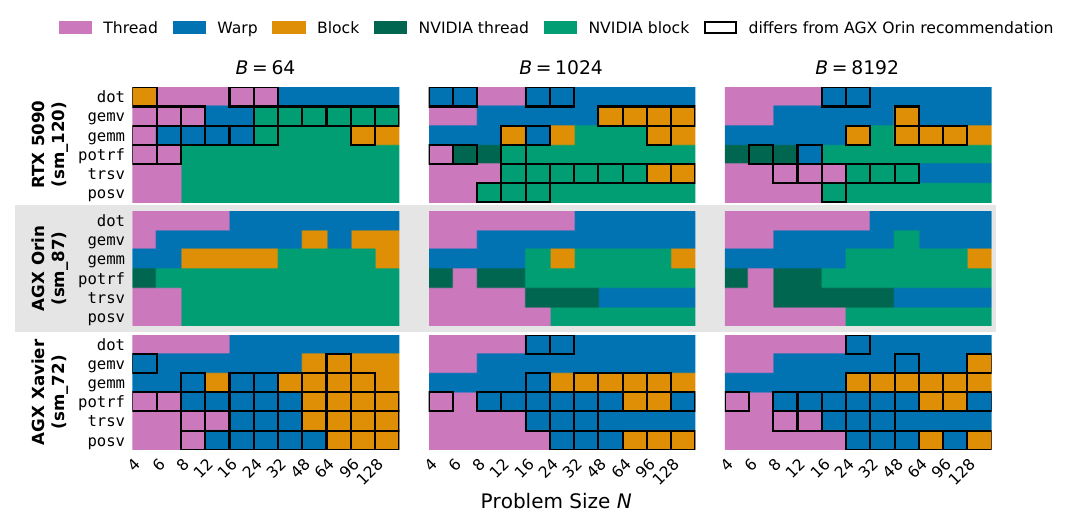}
\vspace{-25pt}
\caption{Recommended \texttt{fp32} placements across operation, problem size, and batch size on the RTX~5090, Jetson AGX Orin, and Jetson AGX Xavier. Colors denote native Thread, Warp, and Block execution and NVIDIA-backed implementations. Black outlines mark recommendations that differ from the Orin's (center row, shaded). For the Xavier much of this divergence is structural as it cannot support the evaluated NVIDIA device libraries.}
\vspace{-10pt}
\label{fig:placement_map}
\end{figure*}

Finally, we note that the NVIDIA warp tier is omitted from Figure~\ref{fig:device_vendor}, and the remaining experiments, as the NVIDIA warp backend is only exposed via CUB reductions, and those fall within the dispatch margin in 121 of 126 cells (96\%).

\subsection{Architecture Adaptation}
\label{sec:placement_eval}

The preceding results show that execution placement matters on one GPU. We next ask whether the same choices transfer across architectures and whether architectural changes impact placement decisions.
Figure~\ref{fig:placement_map} shows that the qualitative trends presented in Section~\ref{sec:device_eval} persist, but their boundaries move substantially. Threads remain particularly effective for many small factorizations and solves, while warp, block, and NVIDIA-backed implementations take over in different regions as problem size and batch size change, reinforcing the importance of a flexible, portable, and architecture-tuned numerical library.

In particular, we find that the Orin's placements differ from the RTX~5090 in 145/396 cells and from the older-generation Xavier in 162/396. 
As the Xavier cannot support the NVIDIA backend, we also restricted all three GPUs to their common native GLASS backend candidates and find that 107/396 RTX--Orin, 94/396 RTX--Xavier, and 66/396 Orin--Xavier placements still change.

\begin{figure}[!t]
\centering
\includegraphics[width=\columnwidth]
{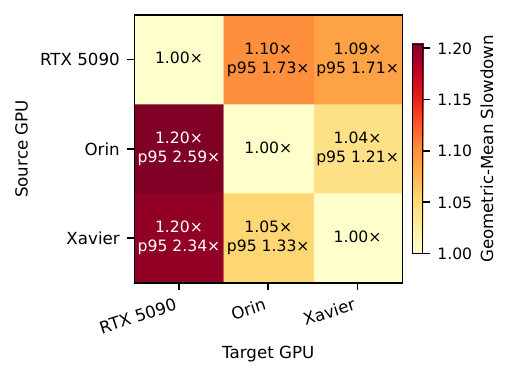}
\vspace{-25pt}
\caption{Penalty from transferring native placements and launch configurations between GPU architectures across the 396 measured cells. Rows are source hardware and columns are target hardware. Each off-diagonal entry reports geometric-mean and 95th-percentile slowdown relative to the locally tuned target plan. One Orin placement (16 warp-packed \texttt{fp64} GEMM problems per block) exceeds the RTX~5090's per-block shared memory and cannot be transferred.}
\label{fig:policy_transfer}
\vspace{-15pt}
\end{figure}

These differences are important for performance. Figure~\ref{fig:policy_transfer} transfers native GLASS placements between architectures. We find that cross-architecture reuse adds 4--20\% geometric-mean runtime depending on direction, with 95th-percentile penalties reaching 2.59$\times$. This is particularly costly when transferring between hardware classes as transfers between the two Jetsons cost only 4--5\%, while carrying a Jetson policy onto the RTX 5090 costs 20\% on average. We note that here, portability means retaining the same numerical interface while retuning placement across NVIDIA GPUs. Cross-vendor source portability is left for future work.

\begin{figure*}[t]
\centering
\includegraphics[width=0.99\textwidth]{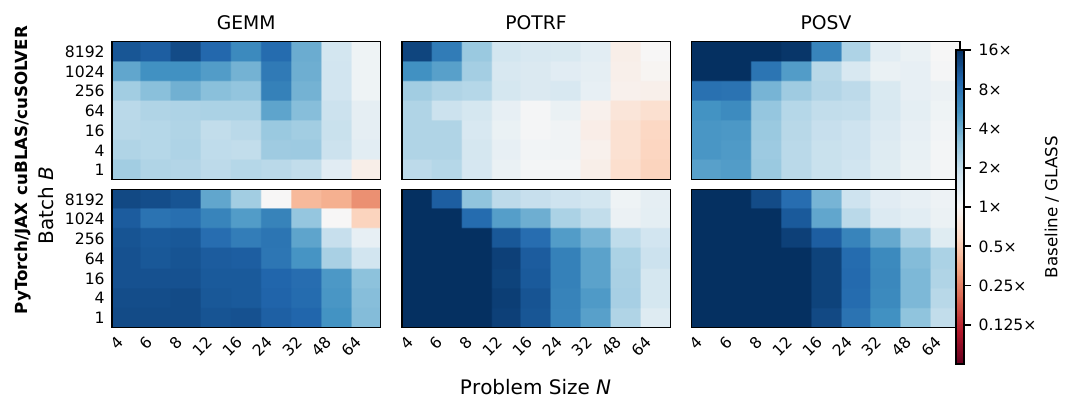}
\vspace{-15pt}
\caption{Host-batched baseline time divided by the best measured valid native GLASS time for \texttt{fp32} GEMM, POTRF, and POSV on the Jetson AGX Orin against cuBLAS/cuSOLVER (top) and the per-cell best of PyTorch and JAX batched linear algebra with device-resident inputs and outputs and pre-compiled and jitted code (bottom). Values above one favor GLASS and are clipped to 16$\times$ for visual clarity (true maximum value is 113.9$\times$), exposing both the small-matrix region where device-side execution wins and the larger regimes where host batching can regain some advantage, and exposing the benefit of a flexible, portable, and adaptable placement framework.}
\label{fig:host_performance}
\vspace{-15pt}
\end{figure*}

For edge deployments, the power mode is also a key consideration. Reducing the Jetson Orin from 50~W to 30~W and 15~W increases geometric-mean runtime by 29\% and 91\%, respectively, yet changes only 9 of 396 placement decisions in each mode (with 5--6 of those sitting inside the dispatch margin). Power limits therefore primarily scale the candidates together, whereas switching architectures changes their relative ordering. The same placement table can consequently be reused across power modes without retuning, even though reducing the power ceiling substantially reduces absolute throughput.

Finally, generating a placement table is a one-time, offline operation for each architecture and toolchain configuration. Our complete tuning sweeps required 5.5 hours on the Orin, 9 hours on the Xavier, and 2 hours on the RTX~5090.

\subsection{Performance Against Host-Batched Libraries and Compiler Frameworks}
\label{sec:perf_eval}

We next compare device-side GLASS execution with conventional host-dispatched numerical libraries and compiler frameworks to demonstrate the overheads associated with such popular approaches. Figure~\ref{fig:host_performance} compares the best measured \texttt{fp32} native GLASS implementation with the best cuBLAS/cuSOLVER or PyTorch/JAX implementation while varying problem size and batch size.

Both baseline families pay per-call dispatch overhead that device-side execution eliminates, and both pay it most heavily in the small-matrix regime GLASS targets. On the Orin, GLASS beats host-dispatched cuBLAS/cuSOLVER at every measured batch size for POSV through $N=64$ (up to 113.9$\times$), GEMM through $N=48$ (up to 12.1$\times$), and POTRF through $N=24$ (up to 12.8$\times$). Compared to the per-cell best of PyTorch and JAX, optimized with both device-resident inputs and outputs and pre-compiled and jitted code, GLASS is faster in 359 of 378 cells with per-operation geometric means of 4.3--9.5$\times$ and individual dispatch-bound cells reaching 73$\times$. That being said, the wins are not uniform, and host cuSOLVER retakes portions of standalone Cholesky at larger $N$, while PyTorch/JAX wins for the largest combination of $N$ and $B$ for GEMM. 

As with the device-side results, ablations show large wins for GLASS on the older Xavier (up to 89.3$\times$ against cuBLAS/cuSOLVER) and smaller wins on the RTX~5090 workstation (GLASS leads the frameworks in 320 of 378 cells at geometric means of 1.5--2.6$\times$, up to 12$\times$). And, unsurprisingly, workstation-only baselines like MAGMA~\cite{haidar2015magma} perform best on the workstation, but remain out of reach for embedded deployments, reinforcing our core thesis that flexible dispatch wins.

\subsection{Robotics Operations and Composition}
\label{sec:robotics_eval}

This placement effect extends to robotics-specific operations and compositions.
Table~\ref{tab:robops} evaluates representative pose, spatial-algebra, small-spectral, and reduction primitives from the robotics-specific portion of GLASS. 
On the Orin, the best fine-grained placement improves throughput by 3.5--9.7$\times$ over whole-block execution across these operations, and the same protocol on the RTX~5090 spans 1.7--5.0$\times$. Thread execution wins most cases, while the reduction-heavy \texttt{fp32} softmax instead favors a warp.

\begin{table}[t]
\caption{Execution placement for representative robotics operations.
ns/problem for \texttt{fp32} on the Jetson AGX Orin
at $B=4{,}096$.}\vspace{-4pt}
\label{tab:robops}
\centering
\small
\begin{tabular}{@{}lrrrr@{}}
\toprule
Operation & Block & Warp & Thread & Block/best \\
\midrule
\texttt{quat\_retract} & 10.291 & 5.752 & \textbf{2.869} & $3.6\times$ \\
\texttt{se3\_retract} & 14.768 & 9.930 & \textbf{4.152} & $3.6\times$ \\
\texttt{quat\_error} & 8.589 & 4.819 & \textbf{2.330} & $3.7\times$ \\
\texttt{motion\_cross\_mul} & 12.092 & 7.322 & \textbf{3.417} & $3.5\times$ \\
\texttt{eig3} & 62.519 & 46.891 & \textbf{8.012} & $7.8\times$ \\
\texttt{svd3} & 77.071 & 56.280 & \textbf{11.831} & $6.5\times$ \\
\texttt{closest\_rotation} & 77.777 & 57.386 & \textbf{8.005} & $9.7\times$ \\
\texttt{argmax (n=16)} & 16.208 & 4.823 & \textbf{1.980} & $8.2\times$ \\
\texttt{softmax (n=16)} & 14.717 & \textbf{4.224} & 11.759 & $3.5\times$ \\
\bottomrule
\end{tabular}

\vspace{-10pt}
\end{table}

\begin{figure}[t]
\centering
\includegraphics[width=\columnwidth]{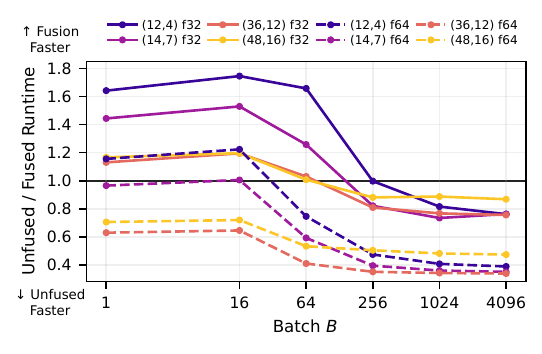}
\vspace{-25pt}
\caption{Runtime ratio of the unfused GLASS Riccati composition (one kernel per primitive, intermediates in global memory) to the fused \texttt{riccati\_gain} kernel (intermediates in shared memory) on the Jetson AGX Orin. Above one favors fusion. Color identifies the problem size, $(n_x,n_u)$. Solid curves are \texttt{fp32}, and dashed \texttt{fp64}.}
\label{fig:riccati}
\vspace{-10pt}
\end{figure}

Composability also makes composition granularity itself a measurable placement choice. Figure~\ref{fig:riccati} evaluates an LQR feedback gain $K = (R + B^\top P B)^{-1} B^\top P A$ expressed two ways from the same GLASS primitives: a fused \texttt{riccati\_gain} kernel that retains every intermediate on chip, and an unfused composition that launches each primitive as its own kernel. Both dominate the equivalent seven-call host-dispatched vendor chain, consistent with Sec.~\ref{sec:perf_eval}, but, as with our prior placement evaluations, the measured winner between them flips with the deployment regime. Through $B=64$ the fused kernel is up to 1.75$\times$ faster, while at larger batches, and for the register-heavier \texttt{fp64} shapes almost everywhere, the unfused composition wins by up to 2.9$\times$. While possibly surprising at first, this is because independent per-primitive kernels can occupy the GPU more fully than the fused kernel's shared-memory footprint allows. Composition granularity is thus another architecture- and regime-specific placement decision GLASS can optimize. %

\subsection{Case Studies in Published Robotics Systems}
\label{sec:casestudies}

We finally integrate GLASS into two existing open-source robotics systems.
Across both, replacing bespoke numerical code preserves or improves performance while reducing application-specific implementation burden, and in one case it also exposes a pre-existing numerical bug.

In a CUDA implementation of sampling-based MPC~\cite{vlahov2024mppi}, we replace application-specific reductions with GLASS, while preserving the surrounding algorithm and public interface. 
This change makes a pre-existing failing host-versus-device normalization test pass. An independent double-precision reference shows that the original reduction can return a non-minimal baseline. Our repair for this error has since been accepted upstream. Relative to the corrected baseline, using GLASS provides 1.21--1.50$\times$ speedups across 128--8192 rollouts on the Orin (1.21--1.46$\times$ on the RTX~5090). Thus, reuse replaces an assumption-sensitive reduction with independently tested primitives while also improving performance.

In a batched inverse-kinematics solver~\cite{yasutake2026hjcdik}, integrating GLASS reduces approximately 355 lines of specialized device numerics to 29 lines of application code, a 92\% reduction. 
For example, a 201-line warp Cholesky implementation becomes \texttt{glass::warp::posv}, while custom reductions, argmin logic, and geometric operations become library calls. 
In an interleaved A/B comparison at batch size 2{,}000, GLASS adoption reduces Orin runtime from 9.94\,ms to 7.92\,ms, a 1.26$\times$ speedup, while the corresponding RTX~5090 improvement is only 1.02$\times$. This is yet another example of why GLASS is useful for edge robotics.

\section{Conclusion and Future Work} \label{sec:conclusion}
GPU robotics repeatedly embeds small, structured numerical operations inside larger kernels, where the best execution strategy depends on the operation, problem size, batch size, and GPU architecture. 
We introduced GLASS, an open source, header-only CUDA C++ library that unifies, under one composable API, thread-, warp-, and block-scoped implementations with NVIDIA device routines, using offline measurement to select among them. 
Our evaluation shows that across three GPU architectures, no implementation family dominates, and transferring placements between architectures adds 4--20\% geometric-mean runtime.
Downstream integrations further show that shared implementations can resolve latent correctness errors while improving embedded runtimes by up to 1.5x.
These results suggest that execution placement should be treated as part of the reusable numerical interface for GPU robotics, rather than repeatedly chosen inside each application.
We release GLASS open source together with its tests, tuning and continuous integration tools, documentation, and coding-agent guidance.

Future work includes provably correct reduced- and mixed-precision numerical support, building on recent tooling infrastructure~\cite{yilmaz2025roboprec}, broader factorization support, community-contributed architecture tables, and support for non-NVIDIA accelerators.
Finally, although motivated by edge robotics, the same design applies wherever small, structured numerical operations are embedded inside larger GPU kernels, and we look forward to supporting other domains.

\ifanon
\else
\section{Acknowledgments}
We used LLM tools, including Codex and Claude Code, to assist with software implementation, drafting, proofreading, and review. All final text, figures, code, and references were edited, revised, and reviewed by humans. We thank Emre Adabag, Miloni Atal, Yana Botvinnik, Sai Coumar, Eric Feng, William Gerard, Brennan McManus, Seyoung Ree, Yang Shaohui, Patarada Yontrarak, and the rest of the A$^2$R lab for their contributions to this codebase.
\fi

\bibliographystyle{styles/IEEEtran_new}
\bibliography{refs}

\end{document}